# On the Handwriting Tasks' Analysis to Detect Fatigue

**Manuel-Vicente Garnacho-Castaño [1], Marcos Faundez-Zanuy [2,\*] and Josep Lopez-Xarbau [2]**

[1] Escola Superior Ciencies de la Salut, Tecnocampus, 08302 Mataró, Barcelona, Spain; mgarnacho@tecnocampus.cat
[2] Escola Superior Politecnica, Tecnocampus, Avda. Ernest Lluch 32, 08302 Mataro, Barcelona, Spain; jlopez@tecnocampus.cat
\* Correspondence: faundez@tecnocampus.cat

**Featured Application: Fatigue detection.**

**Abstract:** Practical determination of physical recovery after intense exercise is a challenging topic that must include mechanical aspects as well as cognitive ones because most of physical sport activities, as well as professional activities (including brain–computer interface-operated systems), require good shape in both of them. This paper presents a new online handwritten database of 20 healthy subjects. The main goal was to study the influence of several physical exercise stimuli in different handwritten tasks and to evaluate the recovery after strenuous exercise. To this aim, they performed different handwritten tasks before and after physical exercise as well as other measurements such as metabolic and mechanical fatigue assessment. Experimental results showed that although a fast mechanical recovery happens and can be measured by lactate concentrations and mechanical fatigue, this is not the case when cognitive effort is required. Handwriting analysis revealed that statistical differences exist on handwriting performance even after lactate concentration and mechanical assessment recovery. (4) Conclusions: This points out a necessity of more recovering time in sport and professional activities than those measured in classic ways.

## 1. Introduction

Handwriting is probably one of the most complex tasks that human beings can perform. It can reveal the identity of the author who wrote a specific text as well as health pathologies such as Parkinson disease (PD) [1] or Alzheimer disease (AD) [2–6]. In this context, it is a simple and non-invasive test for diseases where a biological marker does not exist. It can also be applied as an improvement measurement test after a specific treatment, such as oxygen therapy in patients affected by chronic obstructive pulmonary disease (COPD) [7]. A recent review on online handwriting applications in e-health and e-security is given in [5,8,9]. In [10], one will find a recent view on neurodegenerative dementia.

In this paper we investigated the effects of physical fatigue in handwritten performance of healthy subjects. We proposed a methodology to induce physical fatigue and we analyzed the relation among physical, metabolic, and cognitive fatigue by means of online handwriting analysis.

Fatigue can severely affect sport and professional activity performance, including systems operated by brain–computer interface (BCI), which has been an active research field. In [11], the authors studied BCI systems and investigated how fatigue changes with BCI usage and reported the effect of fatigue on signal quality, analyzing five hours of BCI usage. In [12], the authors investigated the effects of three mental states: Fatigue, frustration, and attention on BCI performance. They found that the relationships among these variables were complex, rather than monotonic, probably due to a poorly induced fatigue, which was not assessed by any other mean. In [13], the authors proposed a novel signal-processing chain inspired from BCI computing to detect mental fatigue.

___________________________________________________________________________________

*1.1. Handwriting Analysis in Fatigue Condition by Calligraphic Experts*

Some works based on visual inspection by calligraphic experts reported the effects of fatigue in offline handwriting, where offline is referred to as the case where the user executes a handwriting task and, later on, the task is analyzed. The book [14] classifies the human stress into two forms: Emotional stress and physical stress. Signatures can be executed by a fatigued writer. For instance, when attending a physical fitness center or signing a receipt. However, it is not usual on more formal documents because of the special circumstances of this act. In these cases it should be enough time to recover from fatigue before writing. This can explain the small number of studies dealing with fatigue and handwriting. First studies were published in [15] (p. 94) and [16] (p. 297), but they are considered as one among several factors that may affect handwriting. In [17] (p. 92) the author reported on the changes in the writing of one subject writing a sentence of eighteen words after having run up four floors of stairs. His findings are summarized in [14]. The author reports changes similar to those produced by intoxication, especially an increase in lateral expansion but without an apparent increase in height. In [18] the authors present a study of 30 writers affected by extreme and moderate states of fatigue and fatigue localized to the writers' forearm. The study evaluated healthy males of very similar age (early twenties) who were asked to write a modified version of the London Letter under four different test conditions. The author observed an increase in vertical height (more than 90% of the analyzed cases) in both lowercase and uppercase letters, without a significant change in proportions or relative heights and an increase in letter width or lateral expansion (77% of the subjects). Considering the spacing between words, the author found both phenomena, expansion and contraction (50% of the subjects), but noted that whichever tendency was exhibited it remained consistent. Slope, speed, rhythm, or fluency habits were not significantly affected. Only minor deterioration was appreciated in writing quality, which tended to produce a scrawl and exhibit less care. In only one case there was greater pen pressure displayed. No evidence of tremor was found. When writing under fatigue, fewer patchings and overwritings occurred. Minute movements tended to be enlarged, but fundamental change to most writing habits was not found. There was some propensity to commit spelling errors, to abbreviate, and to omit punctuation and diacritics ("i" dots). There was no apparent difference between the effects of general body fatigue and forearm fatigue. In either case, however, the severity of the fatigue produced some difference. Roulston's data are fully reported in his work and, while the effects of fatigue cannot be denied, Harrison's statement that "…Fatigue and a poor state of health can have a most deleterious effect upon handwriting…" may be an exaggeration of the condition. In [19], the author presents a study of 21 high school students endeavored to assess whether there was a degree of impairment to a person's writing that would correlate with the pulse rate of the heart under different levels of exertion. This paper studied further what the nature of the impairment would be and whether the author of stress-impaired writing could be correctly identified. The author found that physical stress, producing abnormally high pulse rates, did affect the individual's writing performance, but the reading of pulse rates can only be considered an indicator of the level of stress being experienced by the subject. The author of this study could not conclude that pulse rates are fully responsible for the degree of impairment, although they could be a contributing factor. Impairment of the writing in this case was judged on the strength of the following: (1) Deterioration in letter formation, coupled with overwriting and corrections. (2) An increase in lateral expansion, especially of the spacing between letters, and a frequent misjudgment of the length of words at the ends of lines. (3) A tendency to write larger. (4) A reduction in the speed of writing, accompanied by an inconsistency in point load (pen pressure). (5) A failure to maintain good alignment or a proper baseline. (6) A general failure in writing quality and greater carelessness. A set of 15 experienced writing examiners was able to assign correctly the writing samples of each subject (100 percent accuracy) affected by different levels of stress, despite the impairments evidenced in the writings. These results were consistent with those published in [18]. The author presented an interesting closing remark: The impairment of writing produced by physical stress (after running various distances) was generally similar to, but not as pronounced as, the impairment due

_________________________________________________________________________________

to alcohol ingestion. Thus, extreme fatigue has some effect on the control of the writing instrument, which tends to increase the expansion of the writing both vertically and horizontally. This expansion may be noted in the enlargement of the more minute movements of the writing process and suggests a trend in the writing toward a scrawl. The effects were of relatively short duration and the writing returned to normal when the body had enough time to recover its energy.

These previous-reported studies, found in [14], are based on visual inspection by calligraphic experts rather than computer analyzed documents. Thus, they can be affected by subjectivity of the human examiner.

*1.2. Automatic Fatigue Detection by Computer Analysis*

In this present work we evaluated by computer analysis the effect of physical effort on the skill to perform several online handwriting tasks in healthy subjects. Several published papers deal with the effect of different substances on handwriting: Alcohol [20,21], caffeine [22], and marijuana and alcohol [23]. To the best of our knowledge, this is the first research paper devoted to the study of physical exercise on online handwritten tasks. A deep knowledge of physical exercise influence in handwriting plays a key role in automatic fatigue detection, which is a challenging research topic. Fatigue is an indicator of risk for the occurrence of errors and accidents as well as degraded performance.

Online Handwritten Signal Acquisition

Handwriting signals can be acquired by means of a digitizing tablet with an ink pen (known as online acquisition). This method has the important advantage over the classic method, based on handwriting and posterior scanning (known as offline acquisition), in that the machine can acquire the information "in the air". That is, when there is no contact between pen and paper. Figure 1 shows the acquisition of the word "BIODEGRADABLE" using an Intuos Wacom digitizing tablet (http://www.wacom.eu). The tablet acquires 100 samples per second including the spatial coordinates (x, y), the pressure, and a couple of angles (see Figure 2). The acquired samples are represented by red circles ("o") in the top of Figure 1, and the text can be reconstructed by interpolation between acquired samples (solid line between samples). Next plots represent, respectively, X, Y, pressure, azimuth, and altitude.

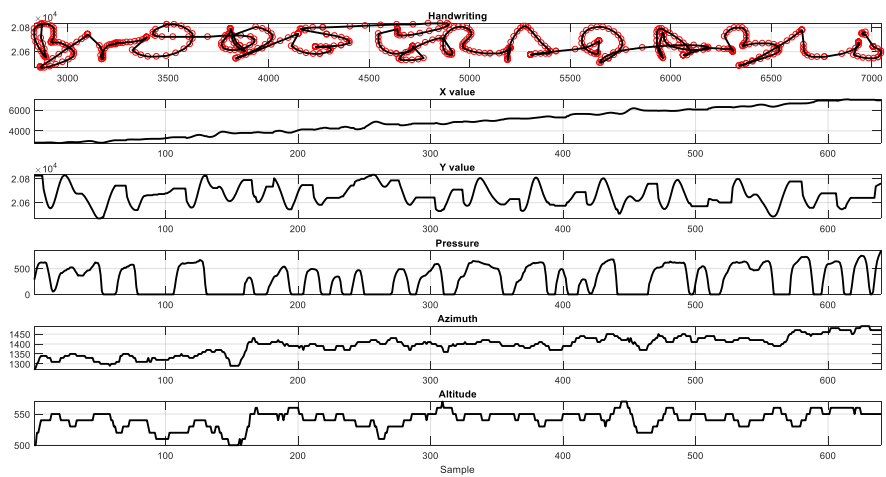

**Figure 1.** Example of handwritten word BIODEGRADABLE input onto a digitizing tablet. Circles represent the acquired samples. *X* axis corresponds to sample number and *Y* axis in the fourth plot to pressure measurement provided by the digitizing tablet, which are nonstandard units.

___

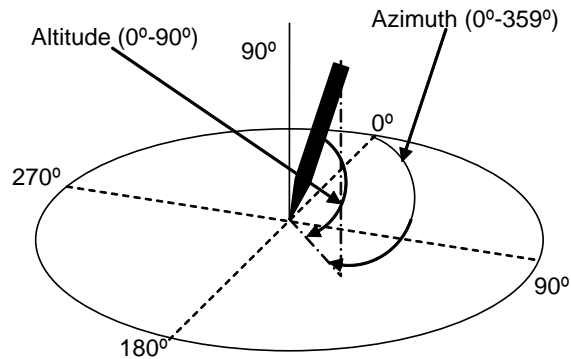

**Figure 2.** Handwriting angle information acquired by the Intuos Wacom (X, Y, pressure, azimuth, altitude).

## 2. Database

In this paper we specially acquired a new handwriting database, described in the next sections. This database will be freely distributed for noncommercial research purposes.

### 2.1. Database Participants

We acquired a new online handwriting database which includes 20 subjects. The selected participants were 20 healthy young men, with the following details:

- Age: 21.3 ± 3.5 years
- Weight: 71.9 ± 7.5 kg
- Height: 175.6 ± 7.2 cm
- Body mass index: 23.2 ± 3.4

All of them were Sport Sciences and Physiotherapy students who performed physical activity a minimum of 3–4 times per week. Participants were familiar with all testing procedures. Exclusion criteria were (1) the use of any medication or performance-enhancing drugs, (2) smoking or alcohol intake, (3) the intake of any nutritional supplement that could alter cardiorespiratory performance, (4) any cardiovascular, metabolic, neurological, pulmonary, or orthopedic disorders that could limit exercise performance, and (5) being an elite athlete. Participants were informed of all experimental tests and signed an informed consent form. The study protocol received approval from the Ethics Committee of the TecnoCampus-Universitat Pompeu Fabra (Mataró, Spain) and adhered to the tenets of the Declaration of Helsinki.

### 2.2. Handwriting Analysis

In several steps of the acquisition protocol a handwriting analysis was performed. Each handwriting analysis consisted of nine different tasks. Figure 3 shows a sample of handwriting analysis with its nine tasks.

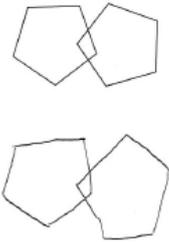

**Figure 3.** Acquisition sheet for each handwriting analysis.

The nine tasks depicted in Figure 4 can be broadly split into three categories:

Mechanical tasks: With no cognitive effort, this task can be performed without any heavy load because it is a repetitive movement that the user can do in an automatic way. He is habituated to do it regularly in his life. This is the case of the handwritten signature (task numbers 4 and 8), handwriting text in capital letters (task number 6), and handwritten text in cursive letters (task 7). Usually, these kinds of tasks are quite straight forward and trivial. These kinds of tasks are quite frequent in the healthy population and find an important niche of applications in biometric recognition of people (user identification and verification [24,25]). Signature acquisition was repeated two times in each recording session to analyze the biometric recognition capability in a future paper.

Cognitive effort tasks: These tasks require some psychical effort to copy a complex drawing. In this case, he requires a strategy to start the task. Some cognitive aspects are important because he needs to know the parts of the drawing already done and the parts that are missing. This kind of task is especially challenging for those people affected by cognitive impairment, such as dementia, mild cognitive impairment, etc. This is the case of the house drawing task (task number 2) and pentagon drawing test (task number 1). Pentagon test is one of the tasks of the Mini-Mental State Examination (MMSE) [26]. The Mini-Mental State Examination (MMSE), or Folstein test, is a brief, 30-point questionnaire test that is used to screen for cognitive impairment. We found significant differences in this kind of task to differentiate mild cognitive impairment from Alzheimer disease in our previous work [2].

Fine motor control tasks: These tasks do not require a heavy cognitive load as the drawing itself is simple and straightforward to understand and memorize at a glance. However, good motor control is required to perform the task. This is the case of the Archimedes spiral drawing test (task number 3), spring drawing task (task number 9), and concentric circle drawing test (task number 5). We found significant differences in this kind of task to analyze Parkinson disease, differentiate essential tremor from Parkinson disease, etc. [1,27,28]. While in some tasks the fine motor control consists of plotting a predefined shape, avoiding the collision with a preprinted drawing (Archimedes spiral), in other

___

cases it requires the skill to perform the movement in a periodic way, keeping on the rhythm (spring and concentric circles tasks).

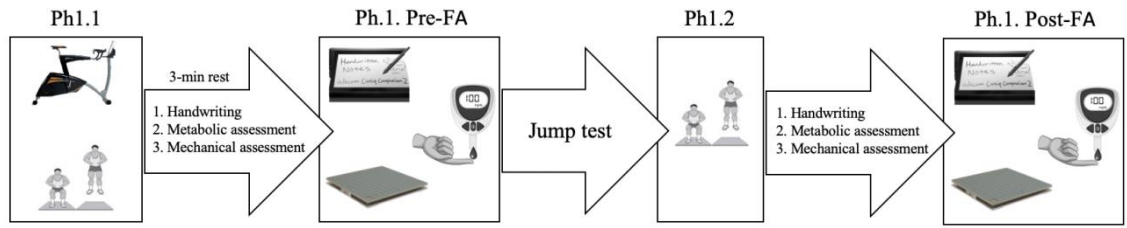

**Figure 4.** Phase 1 sequence of activities. FA = fatigue assessment; Ph = phase.

*2.3. Metabolic Assessment*

Metabolic assessment permits a fatigue estimation. It was measured by blood lactate concentrations. Venous blood (5 μL) was withdrawn by finger pricking by an experienced evaluator. Blood lactate levels were determined using a portable lactate analyzer (commercial model: Lactate Pro 2 LT-1710, Arkray Factory Inc., KDK Corporation, Siga, Japan).

*2.4. Mechanical Assessment*

Mechanical assessment permits a fatigue estimation. Mechanical fatigue of lower limbs was evaluated by countermovement jump test (CMJ) [29,30] on a force platform (Musclelab, Ergotest Technology AS, Langesund, Norway).

Participants carried out three CMJs separated by a rest time of 30 s, and the mean values of vertical flight height and power (3 CMJs) were used in the subsequent analyses. Vertical flight height and power were calculated as follows in equations:

$h = g \cdot f \cdot t^2/8$; h = jump height in meters; g = acceleration due to gravity (m·s$^{-2}$); ft = flight time in seconds. Power (W) = Force (N) · velocity (m/s)

The Borg's scale [31] was used to monitor the rating of perceived exertion (RPE) at the time points established for blood lactate determination.

*2.5. Testing Procedures*

Participants completed one test session at the Exercise Physiology laboratory at our university campus. Sessions were conducted under the same environmental conditions:

- Temperature 20–22.5 °C
- Atmospheric pressure: 745–760 mm Hg
- Relative humidity 40–50%

Participants refrained from any high-intensity physical effort from 48 h and abstained from any type of physical exercise from 24 h before starting the test session. Three test phases were determined.

2.5.1. Phase 1: Jump Test

Fatigue produced by a jump test (JT) was assessed, as indicated in the Figure 4. Phase 1 can be split into four steps as depicted in Figure 5, table 1, and described below:

Ph1.1: A general and specific warm-up was performed. This consisted of a 5 min general warm-up on a magnetically braked cycle ergometer (commercial model: Cardgirus Medical, G&G Innovación, La Bastida, Alava, Spain). It was carried out at a self-selected light intensity, followed by 7 min of dynamic joint mobility drills, stretching exercises, and a specific warm-up (jumps).

PH1.Pre-Fa: Three minutes after finishing subphase Ph1.1, the three assessments were performed: Handwriting, metabolic, and mechanical assessments. These measurements were considered as prefatigue measurements. These three measurements are defined in Sections 2.2, 2.3, and 2.4.

Ph1.2: After 3 min passive rest and assessments, participants performed a JT using the same force platform. JT included three squat jumps (SJ) and three countermovement jumps (CMJ) separated by a rest time of 30 s. The SJ test began from an initial position with hips and knees flexed (~90°) avoiding countermovement and maintaining this position for about 4 s to avoid the buildup of elastic energy stored during flexion to be used by leg extensor muscles. From the position of the hips and knees flexed ~90°, a knee and hip extension was performed as rapidly and explosively as possible. The CMJ test started from a static standing position with hands on their hips. From this position, the subject underwent a rapid flexion-extension of the knees and hips. As in SJ, the depth of the CMJ was determined when the thigh was considered parallel to the floor. The knee joint was at an angle of around 90° of knee flexion.

PH1.Post-Fa: Three minutes after finishing subphase Ph1.2, the three assessments were performed: Handwriting, metabolic, and mechanical assessments. These measurements were considered as post-fatigue measurements.

Fatigue assessment (FA) was performed before (Pre-FA) and after (Post-FA) jump test in Phase 1. FA consists of handwriting acquisition, metabolic, and mechanical analysis.

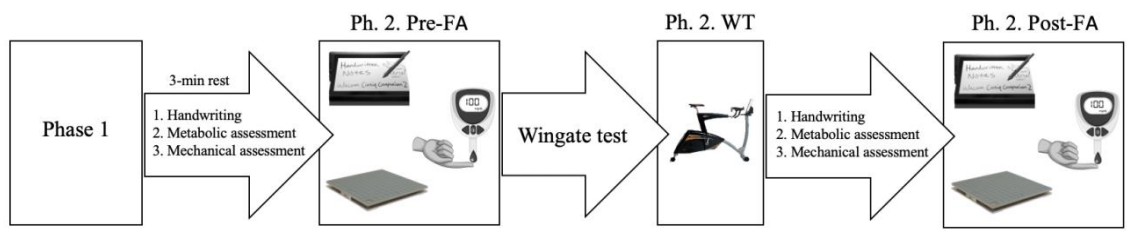

**Figure 5.** Sequence of activities from phase 1 to 2. FA = fatigue assessment; Ph = phase.

2.5.2. Phase 2: Wingate Test

After a 3-min rest (phase 1 completed), a Wingate test (WT) was assessed to determine fatigue induced by lactic anaerobic metabolism [32]. WT is a standard test performed on a stationary cycle-ergometer (previously described) to measure peak anaerobic power and anaerobic capacity. The test consisted of 30 s of cycling at maximum effort with a load (kgf) corresponding to 7.5% of the subject's body weight [33]. The following guidelines were established by the evaluators previous to the Wingate test: (1) In the first seconds, they should pedal from 0 rpm to the highest cadence velocity possible (rpm) in the shortest time possible and (2) maintain the highest cadence velocity possible during the longest time possible until the test end. The same cycle ergometer used in warm-up was individually adjusted to accommodate them to perform a WT. WT was commenced with the subject stopped. Before and after WT, the same procedures were used to assess FA, as previously described (Figure 5)

2.5.3. Phase 3: Fatigue Test

After Phase 2, a 3-min recovery time was provided to participants. Then, immediately fatigue was re-evaluated to determine how participants recovered. As previously, FA was assessed using the same previous procedures, as summarized in Figure 6. In phase 3, no more additional fatigue was induced.

___________________________________________________________________________________________

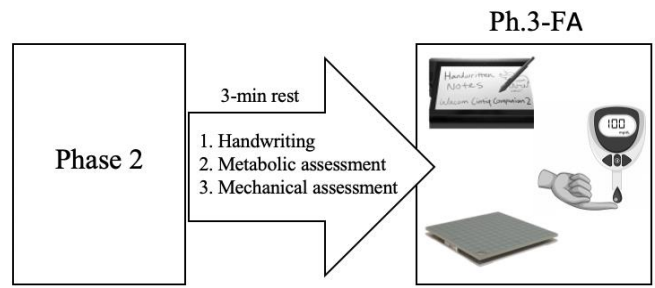

**Figure 6.** Sequence of activities in phase 3 (Ph.3-FA). FA = fatigue assessment; Ph = phase.

As a summary, the following table describes the whole set of measurements:

**Table 1.** Database phases, where S1 to S5 correspond to different acquired set of samples (handwriting, metabolic and mechanical assessments)

| Set | Acquisition Time | Content |
|---|---|---|
| S1 | Ph1-Pre-Fa | Handwriting<br>Metabolic<br>Mechanical |
| S2 | Ph1-Post-Fa | Handwriting<br>Metabolic<br>Mechanical |
| S3 | Ph2-Pre-Fa | Handwriting<br>Metabolic<br>Mechanical |
| S4 | Ph2-Post-Fa | Handwriting<br>Metabolic<br>Mechanical |
| S5 | Ph3-Post-Fa | Handwriting<br>Metabolic<br>Mechanical |

## 3. Experimental Results

### 3.1. Handwriting Feature Extraction

In this paper, we extracted a set of 82 features. From this set, we performed a Wilcoxon statistical test between different phases in order to detect significant differences.

We found significant differences in six features in phase 1/Post-FA, 11 features in phase 2/Pre-FA, 17 features in phase 2/Post-FA, and 29 features in phase 3/FA. Thus, the higher the physical effort, the higher the number of features where significant differences were detected. However, we selected only a subset of these relevant features, which are depicted in Table 2. In this way, we improved the interpretability of results by nontechnical experts (medical doctors, physiotherapists, etc.). The meaning of these features is the following:

Entropy: It is a complexity measure related to the information content. The more unpredictable and complex is a signal, the higher the entropy. The entropy was applied to different signals such as X and Y coordinates, pressure, etc. Entropy is a classic measure used in information theory [34].

Entropy refers to disorder or uncertainty. It is a measure of unpredictability of information content. For instance, the entropy of the pressure signal p, which ranges [0, 2047], can be defined as:

---

$$\text{entropy} = -\sum_{\text{prob}=0}^{\text{prob}=2047} \text{prob} \log_2 \text{prob}$$

where prob is the probability of pressure value p.

In handwriting tasks, entropy is related to the complexity of the writing. The more complex it is, the higher the entropy. This measurement can be applied to trajectories, pressure patterns, etc.

First derivative: The first derivative of a given signal (X, Y coordinate, pressure, etc.) over time is the speed of this signal. Thus, average speed detects if a signal changes fast or slowly. In MATLAB it can be easily calculated. For instance, for the X coordinate:

```
dx = diff(x); %first derivative of x coordinate
```

In handwriting tasks, it is related to the execution speed of the writing. It can be applied to spatial coordinates X and Y and pressure signal, too.

Second derivative: The second derivative of a given signal (X or Y coordinate, pressure, etc.) over time is the acceleration of this signal. Thus, the average acceleration detects the changes of speed of this signal. For instance, for the X coordinate:

```
ddx = diff(x,2); %second derivative of x coordinate
```

In handwriting tasks, it is related to speed changes of the writing. It can be applied to spatial coordinates X and Y and pressure signal, too.

Time in air: Time in air or time up is the time spent with the pen exerting no pressure. This time is considered at short distance (smaller than 1 cm from the tip of the pen to the surface [35]). This time is zero for those tasks where the whole drawing can be produced in a single stroke, and is large when the drawing requires a large amount of strokes.

Given a pressure signal p, it can be obtained with the following MATLAB code:

```
up = find(p == 0);
t_up = length (up) %addition of samples with pressure equal to zero
```

While some movements in air are necessary to change from one stroke to another one, it is also related to pauses and hesitations during the handwriting task. Long times spent in air can be indicative of problems. For example, they are typical in Alzheimer disease.

Normalized time up: The previous feature, time in air, is normalized by the number of strokes in the air. Thus, the previous feature is divided by the number of strokes performed in the air, where a stroke has a specific direction, length, and curvature relative to the other strokes and has been performed between two consecutive instants where the pen touches the surface. Its computation is based on zero-crossing rate (ZCR) value. Zero-crossing rate (ZCR) is the number of times that a given signal crosses from positive to negative (or zero) values or vice versa. It can be defined using the following equations:

```
%zero crossing rate for pressure signal
v = diff(p > 0);
strokes_d = (p(1) > 0) + sum(v == 1); %strokes down
strokes_u = (p(1) == 0) + sum(v == −1); %strokes up
nt_up = t_up/strokes_u; %normalized time up.
```

Time down: Time spent with the pen exerting some pressure. Thus, the tip of the pen is touching the surface of the tablet. The slower the speed of movements, the higher the time down to finish the task.

```
down = find(ne(p,0));
t_down = length(down) %addition of samples with pressure higher than zero.
```

______________________________________________________________________________________

*p* > N: is defined as the number of samples of a specific handwritten task with pressure higher than this N value. Several N values were tested.

pN = sum(p > N).

P[N1-N2]: is defined as the number of samples on a specific handwritten task with pressure between N1 and N2 value, where N1 > 0 and N2 < 1024. Large values imply long time to raise the pressure while small values imply a fast increase in pressure.

Figure 7 shows an example of pressure profile as well as pressure equal to level 100 and 600. The p100 and p600 are defined as the number of samples with pressure value higher than 100 and 600 value, respectively.

Manuel-Vicente Garnacho-Castaño, Marcos Faundez-Zanuy, and Josep Lopez-Xarbau. 2020. "On the Handwriting Tasks' Analysis to Detect Fatigue" *Applied Sciences* 10, no. 21: 7630. https://doi.org/10.3390/app10217630_________________________________________________________________________________

**Table 2.** The $p$ value for several features when applying Wilcoxon test to find differences between sets depicted in Table 1. Initial values at rest correspond to S1. $p$ values smaller than 0.05 are highlighted in bold letters

| | | | Wilcoxon Test $p$ Value When Comparing a Couple of Sets | | | | | | | | | |
|---|---|---|---|---|---|---|---|---|---|---|---|---|
| Task Type | Figure | Feature | S1-S2 | S1-S3 | S1-S4 | S1-S5 | S2-S3 | S2-S4 | S2-S5 | S3-S4 | S3-S5 | S4-S5 |
| Cognitive | 1 | Standard deviation of speed | 0.863 | 0.421 | 0.654 | 0.328 | **0.0412** | 0.1074 | **0.0098** | 0.7717 | 0.4605 | 0.1994 |
| | | Standard deviation of acceleration | 0.84 | 0.421 | 0.579 | 0.328 | **0.0332** | 0.0750 | **0.0098** | 0.7074 | 0.3829 | 0.1994 |
| | | p[100–600] | 0.5 | 0.199 | 0.051 | 0.107 | 0.0985 | **0.0236** | 0.0750 | 0.1858 | 0.3829 | 0.6359 |
| | | Second derivative of pressure | 0.186 | 0.244 | 0.098 | **0.014** | 0.6543 | 0.3641 | 0.0750 | 0.2926 | **0.0370** | 0.0620 |
| | | Max speed | 0.579 | 0.293 | 0.383 | 0.228 | 0.0620 | 0.1168 | **0.0186** | 0.6543 | 0.3829 | 0.2594 |
| | | Max acceleration | 0.598 | 0.31 | 0.346 | 0.244 | 0.0620 | 0.0985 | **0.0164** | 0.6171 | 0.4020 | 0.2594 |
| | | Normalized time up | 0.117 | **0.046** | **0.011** | **0.033** | 0.1858 | 0.1074 | 0.1994 | 0.4213 | 0.6724 | 0.7564 |
| | | Time in air | 0.364 | 0.117 | **0.0458** | 0.075 | 0.3641 | 0.1994 | 0.2283 | 0.2136 | 0.2926 | 0.6543 |
| | | Time down | 0.31 | 0.364 | 0.276 | **0.046** | 0.6171 | 0.3457 | 0.0507 | 0.4408 | 0.0683 | 0.0561 |
| | 2 | Second derivative of pressure | 0.364 | 0.52 | 0.051 | 0.117 | 0.3829 | **0.0209** | 0.0561 | **0.0098** | **0.0186** | 0.6543 |
| | | p > 100 | 0.383 | 0.402 | 0.186 | **0.041** | 0.4802 | 0.1487 | **0.0412** | 0.2136 | 0.0507 | 0.1858 |
| | | p[100–600] | 0.48 | 0.149 | 0.082 | **0.026** | 0.2757 | 0.1074 | 0.0823 | 0.1994 | 0.1487 | 0.5000 |
| | | Time in air | 0.16 | 0.149 | **0.024** | **0.009** | 0.6902 | 0.0620 | **0.0236** | 0.0901 | **0.0412** | 0.2594 |
| | | Time down | 0.293 | 0.383 | 0.186 | **0.019** | 0.5198 | 0.1375 | **0.0209** | 0.1605 | **0.0412** | 0.1994 |
| | | Normalized time up | 0.117 | 0.107 | **0.013** | **9E-04** | 0.4408 | 0.0823 | **0.0209** | 0.1605 | **0.0236** | 0.2926 |
| Mechanical | 6 | Entropy of Y | 0.5 | **0.024** | 0.52 | 0.075 | **0.0370** | 0.6171 | 0.2136 | 0.9814 | 0.8271 | 0.0901 |
| | | Time in air | **0.0412** | **0.011** | **0.002** | **0.002** | 0.0750 | 0.0750 | **0.0458** | 0.4605 | 0.4213 | 0.4213 |
| | | Normalized time up | 0.149 | **0.013** | **0.01** | **0.008** | 0.1268 | 0.0823 | 0.0561 | 0.5000 | 0.4408 | 0.5198 |
| | 8 | Max speed | 0.068 | **0.026** | **0.03** | 0.117 | 0.2283 | 0.5000 | 0.5980 | 0.7406 | 0.7864 | 0.5592 |
| | | Standard deviation of speed | 0.098 | **0.03** | **0.037** | **0.046** | 0.2136 | 0.5980 | 0.5980 | 0.8271 | 0.7717 | 0.5395 |
| Fine motor | 3 | Entropy of X | 0.741 | 0.46 | 0.068 | 0.672 | 0.2136 | **0.0332** | 0.2136 | 0.2926 | 0.6724 | 0.9668 |
| | | p[100-400] | 0.293 | 0.48 | 0.056 | **0.005** | 0.5395 | 0.1074 | **0.0186** | 0.0750 | 0.0507 | 0.3829 |
| | | Second derivative of pressure | **0.0023** | 0.107 | **0.03** | **0.004** | 0.9380 | 0.6359 | 0.4408 | 0.1487 | 0.0750 | 0.2436 |
| | | Time down | 0.1729 | 0.16 | 0.062 | **0.019** | 0.3829 | 0.2436 | 0.0750 | 0.3098 | 0.1074 | 0.2436 |
| | 5 | Max acceleration | 0.974 | 0.383 | 0.801 | 0.772 | **0.0412** | 0.2757 | 0.1168 | 0.9317 | 0.7564 | 0.2757 |
| | | p > 600 | 0.364 | 0.579 | 0.068 | 0.173 | 0.7406 | 0.0507 | 0.1375 | **0.0370** | 0.0901 | 0.6171 |
| | 9 | Standard deviation of acceleration | 0.707 | 0.098 | 0.46 | 0.346 | **0.0458** | 0.4605 | 0.1375 | 0.9099 | 0.7074 | 0.2594 |
| | | Second derivative of pressure | 0.636 | 0.137 | **0.011** | **0.046** | 0.0901 | **0.0332** | **0.0370** | 0.3098 | 0.4408 | 0.5000 |
| | | Time in air | 0.186 | 0.186 | **0.0370** | **0.008** | 0.559 | 0.214 | 0.1074 | 0.127 | 0.068 | 0.46 |

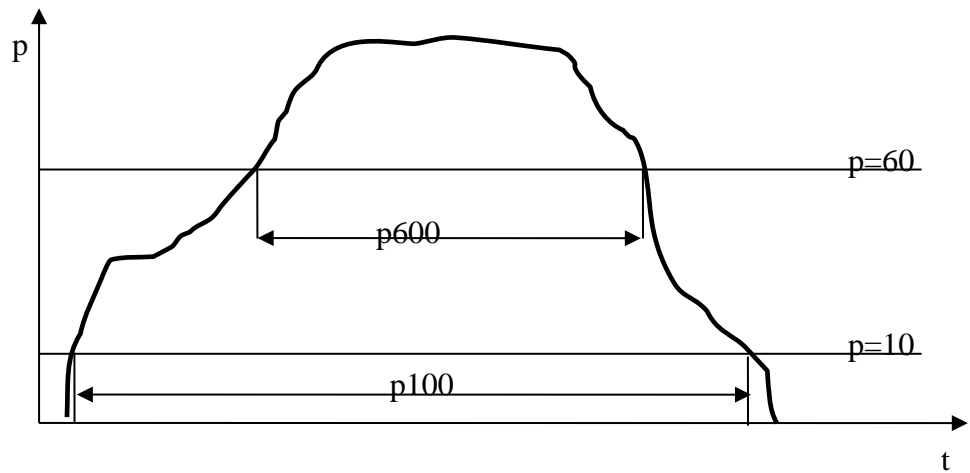

**Figure 7.** Graphical example of p100 and p600.

Max speed: given the instantaneous speed of a signal, its maximum value is selected.

dx = diff(x); %first derivative of x coordinate

dy = diff(y); %first derivative of y coordinate

speed = sqrt(dx.^2 + dy.^2); %instantaneous speed

max_speed = max(speed)

*3.2. Handwriting Experimental Results*

Experimental protocol consisted of extracting several features from handwritten tasks and comparison of phase 1/Post-FA, phase 2/Pre-FA, phase 2/Post-FA, and phase 3/FA with initial values at rest. This comparison was done by means of a Wilcoxon statistical test to find significant differences ($p < 0.05$). Results can be summarized as follows, where a subset of significant features is depicted (Table 2).

*3.3. Metabolic Results*

In blood lactate concentrations, we observed a metabolic x time interaction effect ($p < 0.001$, $\eta p2 = 0.97$, SP = 1.00). A significant metabolic ($p < 0.001$, $\eta p2 = 0.92$, SP = 1.00) and time effect were detected ($p < 0.001$, $\eta p2 = 0.96$, SP = 1.00) (Table 3). Bonferroni multiple comparisons verified lower blood lactate levels in Ph1-Post than in Ph1-Pre ($p = 0.017$; anaerobic alactic metabolism). Higher blood lactate levels were found in Ph2-Post than in Ph2-Pre ($p < 0.001$; anaerobic lactic metabolism). A greater lactate concentration was observed in Ph2-Post than in Ph3 ($p < 0.001$; after recovery time). Also, significant changes were found among Ph1-Post, Ph2-Post, and Ph3 ($p < 0.001$).

**Table 3.** Differences in metabolic and mechanical fatigue and in the ratio of perceived effort between each phase. Data are provided as mean and standard deviation (SD). Abbreviations: ES = effect size; FA = fatigue assessment; Ph = phase; RPE = rating of perceived effort; SP = statistical power. *P*1-values for experimental condition x time interaction effect. *P*2-values for time effect. *P*3-values for experimental condition effect. Bonferroni adjustment: * Significantly different from Pre-FA in each phase (Ph-Pre vs. Ph-Post), (*p* < 0.05); # significantly different from PH.1.Pre-FA and PH.2.Pre-FA, (*p* < 0.05); † significantly different from PH.1.Pre-FA, (*p* < 0.05); § significantly different from PH.1.Post-FA and PH.2.Post-FA, (*p* < 0.001); *f* significantly different from PH.1.Post-FA, (*p* < 0.05); ‡ significantly different from PH.2.Post-FA (*p* < 0.05).

|  | Ph.1. Pre-FA | Ph.1. Post-FA | Ph.2. Pre-FA | Ph.2. Post-FA | Ph.3-FA | *P*1 (ES-SP) | *P*2 (ES-SP) | *P*3 (ES-SP) |
|---|---|---|---|---|---|---|---|---|
| Lactate (mmol L$^{-1}$) | 1.11 (0.17) | 1.01 (0.17) * | 0.99 (0.20) | 14.19 (2.10) *,#,f | 10.49 (2.94) § | <0.001 (0.97–1.00) | <0.001 (0.92–1.00) | <0.001 (0.96–1.00) |
| Flight height (cm) | 36.12 (5.50) | 35.21 (5.47) | 34.86 (5.71) | 33.25 (4.44) *,†,f | 34.09 (4.32) | 0.032 (0.17–0.65) | 0.011 (0.31–0.77) | <0.001 (0.46–0.99) |
| Power output (W kg$^{-1}$) | 29.83 (3.99) | 28.84 (4.29) | 28.59 (4.28) | 27.33 (3.76) *,#,f | 28.37 (3.81) ‡ | 0.003 (0.31–0.91) | 0.060 (0.20–0.48) | <0.001 (0.42–0.99) |
| RPE | 1.13 (1.05) | 1.82 (1.34) * | 1.92 (1.17) † | 8.32 (0.95) *,#,f | 5.26 (1.56) § | <0.001 (0.93–1.00) | <0.001 (0.92–1.00) | <0.001 (0.97–1.00) |

*3.4. Mechanical Results*

In vertical flight height (CMJ test), a significant CMJ x time interaction effect was observed (*p* = 0.03, ηp2 = 0.17, SP = 0.65). A significant CMJ and time effect were detected (*p* < 0.001, ηp2 = 0.46, SP = 0.99; *p* = 0.01, ηp2 = 0.31, SP = 0.77, respectively) (Table 3). Bonferroni assessment confirmed that significant flight height losses were only observed between Ph2-Pre and Ph2-Post (*p* = 0.03; anaerobic lactic metabolism). Also, significant flight height losses were detected only between Ph1-Post (anaerobic alactic metabolism) and Ph2-Post (anaerobic lactic metabolism) (*p* = 0.021).

In power output (CMJ test), a significant CMJ x time interaction effect was verified (*p* = 0.003, ηp2 = 0.31, SP = 0.91). A significant CMJ effect was found (*p* < 0.001, ηp2 = 0.42, SP = 0.99) (Table 3). No time effect was found (*p* > 0.05). Bonferroni post hoc confirmed that significant power output losses were detected between Ph2-Pre and Ph2-Post (*p* = 0.01; anaerobic lactic metabolism) and between Ph2-Post and Ph3. Significant power output losses were found between Ph1-Post and Ph2-Post (*p* = 0.041); however, a greater power output was observed in Ph2-Post than in Ph3 (*p* = 0.02).

Rating of Perceived Exertion (RPE)

A significant RPE x time interaction effect was corroborated (*p* < 0.001, ηp2 = 0.93, SP = 1.00). A significant metabolic (*p* < 0.001, ηp2 = 0.97, SP = 1.00) and time effect were evidenced (*p* < 0.001, ηp2 = 0.92, SP = 1.00) (Table 3). Bonferroni adjustment confirmed lower RPE in Ph1-Pre than in Ph1-Post (*p* < 0.001). A higher RPE was proven in Ph2-Post than in Ph2-Pre (*p* < 0.001) and in Ph2-Post than in Ph3 (*p* < 0.001). Moreover, significant changes were found among Ph1-Post, Ph2-Post, and Ph3 (*p* < 0.001).

**4. Discussion**

Evaluating the lactate concentrations depicted in Table 3, we observed that in phase 1 the lactate concentrations decreased when comparing Ph1.pre versus Ph1. Post. That means there was less fatigue after the first activity (a single jump) than even at rest. However, when looking at Ph.2. and comparing pre and post values, we observed that concentrations increased after pedaling 30 '' as much as they can. In phase 3, it decreased a little bit because the 3 min of recovery were left.

Evaluating the flight height and power in the jump in Table 2, we observed how in the first phase 1 (Ph1.pre vs. Ph1. Post) there were no differences after the first activity because there was no mechanical fatigue and it coincided with the metabolic one. However, in the second phase, there were differences in jump height and power after strenuous activity. Notice in phase 3 how the subjects recovered by increasing power and flight height

In rating of perceived effort, we observed that perception of effort increased dramatically in phase 2 and decreased in phase 3 after recovery.

As expected, we observed that the most intense physical exercise produced more cognitive, metabolic, and mechanical fatigue (S4 being the most intense exercise).

When evaluating the $p$ value for several handwriting features when applying Wilcoxon test to find differences between initial values at rest and different phases, we observed a couple of interesting results:

- Fast fatigue recovery in pure mechanical handwriting tasks, considered as those that do not require too much cognitive effort. This can be observed in Table 2 with the reduction of significant features in Ph3 and the decrease of significance (increase of $p$ values) when compared to Ph1/Pre (S1).
- Slow fatigue recovery in fine motor and cognitive handwritten tasks, considered as those that require high cognitive effort. This can be observed in Table 2 with the increase of significant features in Ph3 and increase of significance (decrease of $p$ values) when compared to Ph1/Pre (S1). In addition, there were no differences between Ph2/post (S4) and Ph3/post (S5), which means no recovery after 3 min rest. It is worth mentioning that in this last phase no differences were found in lactate concentration and mechanical assessment when compared to initial values taken before exercise.
- There were no differences when performing handwriting (signature and cursive letters) or almost no differences (capital letters' task). It is worth mentioning that capital letters' task involves non-usual words, which require some cognitive effort to identify/memorize the word. Probably for this reason, the unique difference was in performance speed.
- There were differences between S3 and S5 in cognitive tasks, especially in task number 2 (house copying test). This implied that cognitive recovering was slower than the mechanical and metabolic one.
- There were almost no differences between S4 and S5. This means that despite the slight recovering of muscular and metabolic fatigue after three minutes of resting, there was no cognitive recovering, which would require more additional time.

## 5. Conclusions

In this paper we proposed a systematic means to analyze fatigue and assess it from three perspectives: Mechanical, metabolic, and cognitive. To this aim, we induced fatigue by means of an intense set of exercises and performed three kind of measurements. The relevance of fatigue is evident in sport and professional activities, including BCI systems. So far, most of the studies rely on single assessment methodology and, to the best of our knowledge, we are the first ones to address this topic from this multimodal perspective (mechanical, metabolic, and cognitive assessment). In addition, the fatigue was induced and measured in a fast way. A new proposed fatigue evaluation was based on comparison on simple handwriting tasks, which can be done in a few seconds.

An important conclusion of this study is that cognitive fatigue requires more recovering time than physical and metabolic fatigue. Thus, physical and metabolic assessments cannot evaluate accurately the required recovering time after strenuous exercise. This time can be evaluated using online handwriting and specific tasks, which require cognitive effort. To this aim, we proposed and evaluated a set of tasks (drawings, handwriting in cursive/capital letters, and signature) and categories of tasks (cognitive, mechanical, and fine motor tasks). Future research will explore a range of innovative, cognitively inspired approaches for contextual unimodal and multimodal processing [36–42].

V.G.-C. and M.F.-Z.; writing—review and editing, M.F.-Z.; funding acquisition, M.-V.G.-C. and M.F.-Z. All authors have read and agreed to the published version of the manuscript.

**Funding:** This research was partially supported by MINECO and FEDER TEC2016-77791-C4-2-R.

**Conflicts of Interest:** The authors declare no conflict of interest.

**Ethical Standards:** All procedures performed in studies involving human participants were in accordance with the ethical standards of the institutional and/or national research committee and with the 1964 Helsinki declaration and its later amendments or comparable ethical standards.